\pdfoutput=1
\documentclass[11pt]{article}
\usepackage[final]{acl}
\usepackage{times}
\usepackage{latexsym}
\usepackage[T1]{fontenc}
\usepackage[utf8]{inputenc}
\usepackage{microtype}
\usepackage{inconsolata}
\usepackage{comment}
\usepackage{tcolorbox}
\usepackage{color,soul}
\usepackage{MnSymbol}
\usepackage{cleveref}
\usepackage{adjustbox}
\usepackage{tabularray}
\usepackage{hhline}
\usepackage{enumitem}
\usepackage{changepage,threeparttable}
\usepackage{longtable} 
\usepackage{booktabs}  
\usepackage{caption}
\usepackage{float}
\usepackage{csquotes}

\usepackage{hyperref}
\definecolor{myblue}{RGB}{86,156 , 214}
\usepackage{mathtools}

\newcommand{\sys}{\emph{TRATES}}

\title{\sys: Trait-Specific Rubric-Assisted Cross-Prompt Essay Scoring
}

\author{Sohaila Eltanbouly, Salam Albatarni, Tamer Elsayed \\
Computer Science and Engineering Department, Qatar University, Doha, Qatar\\
         \{se1403101, sa1800633, telsayed\}@qu.edu.qa\\
         }

\usepackage{todonotes} 

\usepackage{tabularray}
\usepackage{amsmath}

\begin{document}
\maketitle
\begin{abstract}
Research on holistic Automated Essay Scoring (AES) is long-dated; yet, there is a notable lack of attention for assessing essays according to individual traits. In this work, we propose \sys, a novel trait-specific and rubric-based cross-prompt AES framework that is generic yet specific to the underlying trait. The framework leverages a Large Language Model (LLM) that utilizes the trait grading rubrics to generate trait-specific features (represented by assessment questions), then assesses those features given an essay. The trait-specific features are eventually combined with generic writing-quality and prompt-specific features to train a simple classical regression model that predicts trait scores of essays from an unseen prompt. Experiments show that \sys{} achieves a new state-of-the-art performance across all traits on a widely-used dataset, with the generated LLM-based features being the most significant.

\end{abstract}

\section{Introduction} 
Automated Essay Scoring (AES), stemming from Page's early study~\cite{page1966imminence}, has seen notable progress in addressing writing evaluation. It covers holistic scoring (giving a single score for overall proficiency) and trait scoring (assessing specific writing aspects like organization). While holistic scoring offers a broad assessment of writing ability, trait scoring provides detailed feedback to aid students in targeted skill improvement. Due to the complexity of evaluating different traits, holistic scoring has been the predominant focus in AES research. Various methodologies have been introduced, ranging from approaches relying on hand-crafted features~\cite{phandi2015flexible} to language model-based ones \cite{xie2022automated}. 


Cross-prompt AES is gaining momentum in recent AES research, with the goal of training a model that can effectively score essays from \emph{unseen} prompts. 
This approach is not only more practical, reflecting real-world scenarios where models must generalize across diverse prompts, but also more challenging, as it demands the ability to adapt to variations in writing styles, topics, and prompt structures with high accuracy. Different approaches have been proposed for cross-prompt AES, including feature-based approaches \cite{li-ng-2024-conundrums}, multi-task learning \cite{ridley2021automated}, and
contrastive learning \cite{chen-li-2024-plaes,chen-li-2023-pmaes}.


As Large Language Models (LLMs) represent the recent advancements in NLP,  
there is a growing trend towards the integration of LLMs in AES \cite{naismith-etal-2023-automated,do-etal-2024-autoregressive}.  
The common approach involves providing the LLM with task descriptions, rubrics, and essays for scoring. However, this method has not achieved the desired results, falling short of basic baseline performance \cite{yancey-etal-2023-rating,mansour-etal-2024-large}. Other studies employed LLMs as conversational models, where scoring is performed on multiple steps \cite{stahl-etal-2024-exploring, lee-etal-2024-unleashing}. 
While these approaches improve upon basic LLM prompting, they still lag behind baseline performance. This highlights the need for a hybrid approach integrating the advanced text analysis capabilities of LLMs with the previously well-established AES methods. 

In this work, we address the challenge of \emph{trait-based cross-prompt AES} with a novel approach that redefines the role of LLMs. Rather than following the conventional paradigm of ``\emph{Given this essay, provide a score.}'', 
we propose a hybrid framework, \textbf{\sys},\footnote{Spelled differently from `Traits' with same pronunciation.} a Trait-specific and Rubric-Assisted Cross-Prompt AES framework, that leverages generic writing quality features in addition to \emph{trait-specific} features, automatically generated via an LLM, that are easy to interpret, thus providing direct feedback to students on sub-trait aspects.

At its core, \sys{} leverages an LLM to generate and extract trait-specific features given a trait rubric in two stages. Initially, a set of questions is generated from the rubric to assess a specific trait. The LLM then answers each question individually, given the essay. 
The intuition is to streamline 
the rubric to facilitate the assessment of various aspects within the trait, rather than assessing the trait as a whole. 
Finally, trait-specific features are combined with writing-quality and prompt-specific features to train a cross-prompt classical \emph{regression} model that predicts the trait scores of essays from unseen prompts.
This approach allows us to handle the scoring of various traits within a unified framework while automatically tailoring rubric-based features for each specific trait, a task that would be difficult to achieve without the capabilities of LLMs. 
Our contribution is five-fold: 
\begin{enumerate}[itemsep=0mm]
    \item We present \sys, a \emph{novel} rubric-assisted framework for trait-based cross-prompt AES that provides a simple and generic pipeline that can be used to score any trait while generating features specific to the trait. It combines the strengths of powerful LLMs (for feature generation and extraction) with the simplicity of a basic regression model (for scoring).
    \item  \sys{} establishes a new state-of-the-art (SOTA) performance on \emph{all} traits on a widely-used dataset.
    \item We conduct an ablation study to show the significance of each feature category.
    \item We assess the generalizability of \sys{} over two different datasets.
    \item We publicly release all trait-specific features to enable future research.\footnote{\url{https://github.com/Sohaila-se/TRATES}}
\end{enumerate}

The remainder of this paper is organized as follows. Section \ref{sec:related-work} outlines the related work. Section \ref{sec:cross-prompt} defines the cross-prompt AES problem.
Section \ref{sec:trates_framework} details our proposed \sys{} framework. Section \ref{sec:exp_setup} discusses our experimental setup. 
Section \ref{sec:results} presents the results and offers a comprehensive analysis. Finally, Section \ref{sec:conclusion} concludes with few suggested future work directions.

\section{Related Work} 
\label{sec:related-work}
In this section, we review AES studies, focusing on cross-prompt trait scoring and LLM integration.

\paragraph{Cross-Prompt AES}

Early cross-prompt studies concentrated on holistic scoring \cite{jin-etal-2018-tdnn, LI2020106491, ridley2020prompt}, employing methods ranged from simple neural networks to hybrid models combining neural networks with engineered features. However, holistic scoring falls short in providing detailed feedback. To address this, \citet{ridley2021automated} introduced a new task: cross-prompt trait scoring, and developed a POS-embedding-based neural model. Building on this, \citet{do-etal-2023-prompt} enhanced the architecture by incorporating prompt-text features, achieving SOTA performance on the ASAP dataset. Several learning methods have been applied to cross-prompt trait scoring, including multi-task learning \cite{li-ng-2024-conundrums}, contrastive learning \cite{chen-li-2023-pmaes}, and meta-learning \cite{chen-li-2024-plaes}. 

\paragraph{LLM for AES}

Research on LLMs for AES is expanding but remains limited. Promising results were demonstrated by \citet{do-etal-2024-autoregressive} with fine-tuning T5 for prompt-specific scoring. Aside from fine-tuning, several LLM prompting techniques were applied to AES~\cite{mansour-etal-2024-large}. GPT-4 showed improvement with few-shot examples; 
however, it failed to outperform a simple XGBoost baseline \cite{yancey-etal-2023-rating}. Similarly, GPT-3.5, used for scoring and feedback generation, was unsuccessful
\cite{han2023fabric}. Moreover, \citet{stahl-etal-2024-exploring} explored various LLM prompting strategies for scoring and feedback generation, including impersonation and chain of thought. Recently, \citet{lee-etal-2024-unleashing} proposed a Multi Trait Specialization framework that engages the LLM in a conversation to score the essay holistically. 

Most of the proposed LLM-based approaches above focused on zero-shot holistic essay scoring, raising multiple concerns regarding inconsistent evaluation and hallucinations. In contrast, \sys{} introduces a novel use of LLMs, serving as feature generators and extractors, rather than direct graders.
It then trains a regression model using LLM outputs (besides other features), resulting in more effective scoring while eliminating the need for direct scoring with the LLM, which has been shown to be ineffective. Additionally, while most of the work targets holistic scoring, our framework focuses on trait scoring and ensures a more comprehensive evaluation.
Moreover, unlike \sys, none of the related work that utilized LLMs has outperformed or even reached SOTA performance. 
Finally, this work addresses key gaps in cross-prompt trait scoring; unlike prior approaches, our framework integrates the scoring rubric to extract trait-specific features that are easily interpretable. 

\begin{figure*}[ht] 
     \centering
 \includegraphics[width=\textwidth]{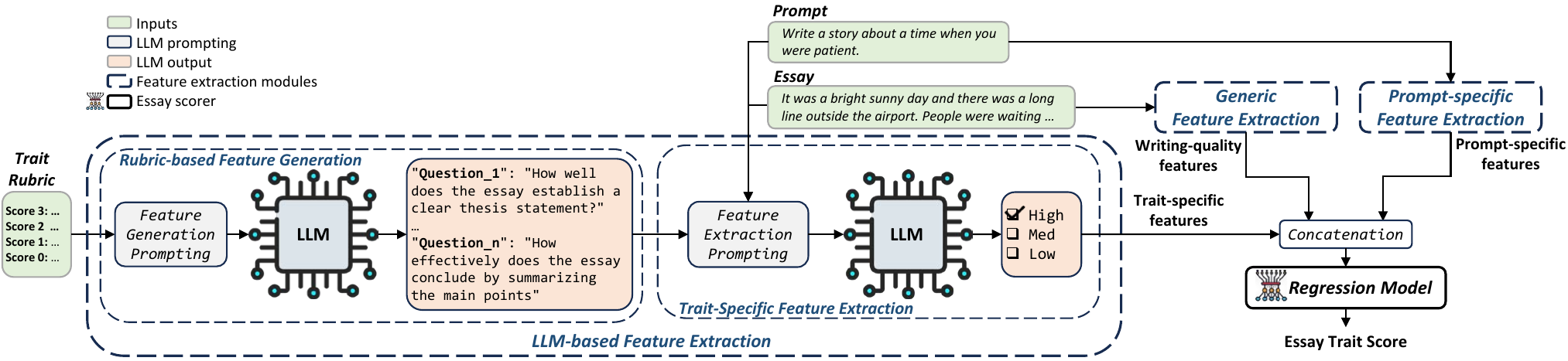}
     \caption{Overview of our \sys{} framework.}
     \label{fig:high-level-arch}
 \end{figure*}
 
\section{Cross-prompt Trait-based AES} \label{sec:cross-prompt}
In trait-based AES, a prompt $p$ is defined as a tuple ($a_p$, $T_p$, $E_p$), where $a_p$ is the prompt task description (a paragraph or so describing the writing task), $T_p$ is a set $\{(t, r_t)\}$ of traits; each trait $t$ (an aspect of student writing, such as organization or sentence fluency) is associated with a rubric $r_t$ (a set of criteria used to evaluate the specific trait $t$), and $E_p$ is a set 
$\{(e,\{s_{e,t}\})\}$ of essays written for the prompt $p$; each essay $e$ is associated with a score $s_{e, t}$ for each trait $t \in T_p$. While, generally, each prompt has its own trait rubrics, those rubrics are usually common across different prompts for specific traits.

The cross-prompt problem is set up as follows. We aim to build a model that is trained on a set of source prompts $P_{src}$ to score the (traits of) essays written for a target prompt $p_{trg} \notin P_{src}$. We note that only the task description of the target prompt is available to the model at inference time. 



\section{\sys{} Framework} \label{sec:trates_framework}

In essay scoring, the rubric generally serves as a scoring guide that sets out the criteria for different performance levels. For example, a brief 3-level rubric for the \emph{organization} trait could be: 
\begin{quote}
    \textit{\textbf{Score 2:} Organization and connections between ideas are logically sequenced.} \\
    \textit{\textbf{Score 1:} Organization and connections between ideas are weak.} \\
    \textit{\textbf{Score 0:} No organization evident.}
\end{quote}
Human evaluators, typically teachers, use the rubric to assess the student essays, which enhances grading consistency and transparency. Moreover, it provides students with detailed feedback pinpointing areas for improvement.  
Given the rubric's importance in \emph{manual} essay scoring, our primary objective is to develop an \emph{automated} rubric-based scoring framework that offers the same benefits as manual scoring, focusing on assessing essay traits.

The main challenge in automatically scoring essay traits lies in identifying the essay representation that best reflects the characteristics of each trait.
To address that, we propose \sys, a Trait-specific and Rubric-Assisted AES framework, illustrated in Figure \ref{fig:high-level-arch}. \sys{} leverages the rubric to identify \emph{trait-specific} features (represented by automatically-generated questions), then extracts the values of those features (by answering the generated questions) from the essays, using an LLM. 
Those features combined with prompt-specific and generic features are used to train a (relatively-simple) regression model for scoring essay traits. 

The major advantage of \sys{} is that it is generic enough to be applied to \emph{any} trait, given the rubric used to assess that trait, while being able to generate and extract different features that are more specific to that trait. 
In this section, we discuss the main components of the framework in detail.

\subsection{Rubric-based Feature Generation} 
\label{sec:questions_generation}

The first component aims to automatically generate (or identify) essay features that are trait-specific and rubric-based. To achieve that, we use an LLM to convert a given trait rubric into a set of assessment questions (acting as \emph{sub-traits}) that constitute trait-specific features. This helps assess the various individual aspects associated with each trait, rather than considering all the aspects in one assessment step, allowing for a more fine-grained assessment of the essay traits. 

To generate the trait-specific features $Q_t$ for a trait $t$, an LLM $\mathcal{M}$ is prompted, with LLM prompt instructions $I_{gen}$, to formulate a set of assessment questions $Q_t$ based on the given trait rubric $r_t$: 
\begin{equation}
    Q_t = \mathcal{M}(r_t | I_{gen})
\end{equation}
where $Q_{t} = \{q_1,q_2,...,q_n\}$ is the set of $n$ sub-traits generated from $r_t$. The LLM is prompted to formulate the questions to rate the essay's aspects as \emph{high}, \emph{medium}, or \emph{low}. 
The rationale is (1) we opt for simple-to-assess questions, (2) we avoid numeric ratings because the generated questions may not always be formulated so that higher ratings imply better quality, and (3) we prioritize clear and easily understandable responses.
The LLM-prompt is shown in Figure \ref{fig:question_generation_prompt}. We note that the same LLM-prompt template is used with \emph{any} LLM and trait, with adjustments made to include the specific essay type (if any), trait name, and rubric (as is). 


\subsection{Trait-specific Feature Extraction} 
\label{sec:feature_extraction}

We next use the same LLM $\mathcal{M}$ to answer each sub-trait question $q_i \in Q_{t}$, given the LLM prompt instructions $I_{ans}$ (see Appendix \ref{appendix:llms_prompts}) 
and essay $e$. 
\begin{equation}
    v(e, q_i) = \mathcal{M}(e, q_i | I_{ans}) 
\end{equation}
where $v(q_i) \in \{\text{high}, \text{medium}, \text{low}\}$ is the answer (i.e., feature value) to the question $q_i$ for essay $e$. 

\subsection{Prompt-Specific Feature Extraction}
\label{sec:prompt_feature_extraction}
Training data for cross-prompt AES typically involves essays of various types, such as persuasive and narrative. The writing requirements for each type can differ significantly; for instance, persuasive essays must be supported by arguments, whereas narrative essays often rely on creative writing. Additionally, essays from different prompts may vary in length depending on the specific writing requirements and the grade level of the students, which can greatly impact the assigned score. 

Since different prompts exhibit distinct characteristics, we define a set $O$ of \emph{prompt-specific} features 
that represents properties of the writing prompts, including 
\emph{essay type} (e.g., persuasive or narrative), \emph{expected essay length} (aligned with task requirements), \emph{source length} (for tasks that utilize external sources), and the \emph{grade level} of the students. The rationale behind these features is to enhance the model’s ability to distinguish between different prompts and to establish a connection between prompts with similar characteristics, indicating to the regression model that these prompts share certain writing characteristics.

\subsection{Generic Feature Extraction}
\label{sec:generic_feature_extraction}
While the LLM-based features are trait-specific, there are other features that are \emph{generic} that can help capture different aspects of writing proficiency. 
We considered five categories of features that were widely utilized in the literature \cite{ridley2020prompt,ridley2021automated}:
(i) \emph{length-based} features ($G_L$), such as the number of words and sentences in the essay, 
(ii) \emph{readability} ($G_D$) features, which measure how difficult the essay is to read, (iii) \emph{text variations} ($G_T$) features covering the usage of different part-of-speech tags (POS) and punctuation, 
(iv) \emph{text complexity} ($G_C$) features, which evaluate the structural complexity of essays, and 
(v) \emph{sentiment} ($G_S$) features assessing the tone of the essays. This feature set is denoted by $G = \{G_L, G_D, G_T, G_C, G_S\}$. 
The full feature list $G$ is presented in Appendix \ref{appendix:wq_features}. 

\subsection{Trait Scoring}
\label{sec:regression}

At the final stage of the framework, the concatenated list of extracted values of the trait-specific $Q_t$, prompt-specific $O$, and writing quality $G$ features from each essay of the set of source prompts $P_{src}$ are used to train a cross-prompt and trait-specific regression model $R(t)$, which is then used to predict the scores of essays from an unseen prompt $p_{trg}$. 
Note that the feature sets $O$ and $G$ are common for all traits, while $Q_{t}$ differs across traits, resulting in a trained model for each trait.

We use regression models instead of classifiers to provide our model with the ability to predict the essay score with granularity similar to real-world scenarios. We opted for a \emph{shallow neural network} to maintain simplicity in the regression model and to account for the expected relatively small amount of training data.

\subsection{Addressing Cross-prompt Challenges}
\label{sec:cp-challenges}

The main challenge in cross-prompt AES lies in integrating various writing tasks with different characteristics and requirements into a single scoring model; differences between those tasks can manifest at various levels, such as variations in essay types that have distinct writing requirements, different scoring rubrics with unique criteria and score ranges, and the varying quality of writing expected across grade levels.
To ensure the model's generalizability to unseen prompts, it must account for potential differences that may arise during inference. 
In this section, we discuss how \sys{} attempts to address these challenges.

\paragraph{Different scoring rubrics} Having different essay types implies having different rubrics. This imposes a challenge because what qualifies as a strong essay under one rubric may differ significantly from another. Given that our methodology relies on the scoring rubric, we utilized the rubrics from all source prompts for the LLM-based feature extraction, assuming that their diversity would cover the scoring criteria of different essay types. 

\paragraph{Different score ranges} One challenge with using prompts with different rubrics is that each may have a different scoring range. This poses a challenge when training a regression model, as it requires scores to be standardized within a unified range. A direct scaling method is to apply min-max scaling within each prompt, ensuring that all score ranges are mapped to a common scale. 
However, this ignores grade level differences, where the quality required for a maximum score varies. For example, an 8$^{th}$-grade essay might earn a maximum score for clear ideas and basic structure, while a 12$^{th}$-grade essay requires sophisticated analysis and advanced writing for the same score. A min-max scaling would give the same unified score for both.
To address this issue, we propose a score-scaling method based on incremental adjustments relative to the highest grade level; scores for a given grade level are scaled within a fixed range, where the minimum score remains constant while the maximum score decreases by one level for each grade below the highest. This ensures the scaled scores are accurately comparable across prompts with different grade levels and rubrics.

\section{Experimental Setup} \label{sec:exp_setup}
In this section, we outline the setup used to conduct our experiments, including the datasets, selected LLMs, baselines, and implementation details.

\paragraph{Datasets} 
In our main experiments, we used the Automated Student’s Assessment Prize (ASAP)\footnote{\url{https://www.kaggle.com/c/asap-aes}} and ASAP++~\cite{mathias2018asap++} datasets combined, which are widely used for AES evaluation. ASAP has 8 prompts (P1-P8) with trait annotations for P7 and P8 only; ASAP++ extends it by scoring traits for P1-P6. Table \ref{tab:dataset} describes the dataset. The traits are: Content (CNT), Organization (ORG), Word Choice (WC), Sentence Fluency (SF), Conventions (CNV), Prompt Adherence (PA), Language (LNG), and Narrativity (NAR). 

To test the generalizability of \sys, we conducted additional experiments over the ELLIPSE dataset \cite{crossley2023english} (Table \ref{tab:dataset_ellipse}), which comprises about 6.5k essays written by English Language Learners for 44 prompts. 
Each essay is assessed over 6 traits (cohesion (COH), syntax (SYN), vocabulary (VOC), phraseology (PHR), grammar (GRM), and conventions (CNV)) using a standardized rubric with a scoring range [1-5] and increments of 0.5 points.

\paragraph{LLMs Selection} 
The LLMs we experimented with are chosen based on 4 criteria: (1) we opt for open-source models for accessibility, reproducibility, and cost-effectiveness, (2) they are based on different foundation models, (3) considering efficiency and resource constraints, we limit our selection to smaller-scale LLMs of size ranging from 7B to 9B parameters, and finally, (4) we consider the highest-ranked models on the Arena Elo benchmark (at the time of experiments) that match the criteria above.\footnote{\url{https://huggingface.co/spaces/lmsys/chatbot-arena-leaderboard}} For each LLM, we used its corresponding checkpoint available on Hugging Face.

Accordingly, we selected three LLMs: (i) \textbf{
Starling}-LM-7B-beta~\cite{starling2023}; (ii) \textbf{Llama}-3.1-8B-Instruct~\cite{touvron2023llama}; and (iii) \textbf{Gemma}-2-9b-it-SimPO~\cite{meng2024simpo}. More details are provided in Appendix \ref{appendix:llms_desc}. 
\paragraph{Baselines}
To evaluate \sys, we compare it with 3 baselines. The first two are considered the \emph{SOTA} for cross-prompt AES on the ASAP dataset.

\setlist{nolistsep}
\begin{itemize}[noitemsep]
    
    \item \textbf{ProTACT} \cite{do-etal-2023-prompt} a cross-prompt model that leverages prompt attention, topic-coherence features, and trait-similarity loss.
    \item \textbf{Li \& Ng} \cite{li-ng-2024-conundrums} adopted a simple neural architecture with different sets of features, developing feature-based models.
    \item Zero-shot LLM (\textbf{LLM-D}) utilizes an LLM to \emph{directly} scores essays with zero-shot prompt. 
    The details are presented in Appendix \ref{appendix:zs_llm_scoring}.
\end{itemize}

For evaluation, we use Quadratic Weighted Kappa (QWK)~\cite{cohen1968weighted}, a common measure for AES that assesses the agreement between the scores of two raters.


\paragraph{Training and Hyperparameter tuning} For ASAP, we used leave-one-prompt-out cross-validation. The hyperparameters of the models are tuned using sequential hyperparameter tuning, where one hyperparameter is optimized at a time while keeping others fixed. The hyperparameters and their search space are listed in Table \ref{tab:hyperparameters}.
We used early stopping with a patience of 10 epochs, and ReduceLROnPlateau learning-rate scheduler with a patience of 5 epochs and a factor of 0.1. The batch size was fixed to 128 for all the experiments. LLM's answers (high/medium/low) were mapped to numerical values (3/2/1). 
We report the performance of our models on the \emph{unseen target prompts} for each trait and on average. 
The performance of ProTACT and Li \& Ng baselines is reported as in their corresponding studies for ASAP dataset. For ELLIPSE dataset, we trained the ProTACT model using their implementation.\footnote{\url{https://github.com/doheejin/ProTACT}} 

\paragraph{Feature Generation} 
Trait-specific features are generated from each unique rubric.
Tables \ref{tab:stat_question} and \ref{tab:question} show the feature count and examples for ASAP, while Table \ref{tab:question_ELLIPSE} presents the same for ELLIPSE.

\paragraph{Prompt-specific and Generic Features}
The prompt-specific features are extracted from the metadata provided with the dataset.  The generic features, described in Table \ref{appendix:feature_list}, are extracted using \cite{ridley2021automated} code.\footnote{\url{https://github.com/robert1ridley/cross-prompt-trait-scoring} }

\paragraph{Feature Normalization}
Previous studies typically normalize features within individual prompts, assuming that the feature distribution of the unseen prompt is known~\cite{do-etal-2023-prompt,li-ng-2024-conundrums}. However, this assumption does not align with real-world scenarios. Instead, we aim to develop a generalizable model that can handle unseen prompts of any size. To achieve this, we use the minimum and maximum values from the \textit{training} dataset for normalization and consistently apply those values to the test data during inference.

\paragraph{Score-scaling} Due to inconsistency in score ranges of ASAP prompts, we scaled all scores to [0-6]. We also implemented score scaling based on grade levels (Section \ref{sec:cp-challenges}), ensuring the maximum score for lower grade levels is appropriately reduced. More details are provided in Appendix \ref{appendix:score-scaling}. The predicted scores are scaled back to their original range for evaluation to ensure a fair comparison with previous work.  





\section{Results and Discussion} \label{sec:results}
In this section, we discuss the results of our experiments addressing 4 research questions: 
\textbf{RQ1}: Are features generated and extracted via LLMs effective? (\ref{sec:RQ1}), 
\textbf{RQ2}: Can \sys{} with trait-specific and generic features improve the performance? (\ref{sec:RQ2}), 
\textbf{RQ3}: Which feature category holds greater significance? (\ref{sec:RQ3}), and 
\textbf{RQ4}: How well does \sys{} generalize? (\ref{sec:RQ4}). Furthermore, we discuss the inference cost in Section \ref{sec:cost}.

\subsection{Effectiveness of LLM Features (RQ1)} 
\label{sec:RQ1}

\begin{table*}\centering
\setlength{\tabcolsep}{4pt}
\normalsize	
\renewcommand{\arraystretch}{1}
\begin{tabular}{lll|llllllll|l}
\hline
&\textbf{Model} &\textbf{LLM} &\textbf{ORG} &\textbf{WC} &\textbf{SF} &\textbf{PA} &\textbf{NAR} &\textbf{LNG} &\textbf{CNV} &\textbf{CNT} &\textbf{Avg} \\
\hline
a&ProTACT &- &0.518 &0.599 &0.585 &0.619 &0.639 &0.596 &0.450 &0.596 &0.575 \\
b&Li \& Ng &- &0.478 &0.459 &0.452 &0.617 &0.637 &0.556 &0.439 &0.592 &0.529 \\
\hline \hline
c&LLM-D &Starling &0.281 &0.318 &0.290 &0.322 &0.289 &0.221 &0.157 &0.282 &0.270 \\
d&&Llama &0.323 &0.183 &0.209 &0.487 &0.467 &0.484 &0.168 &0.332 &0.332 \\
e&&Gemma &0.345 &0.375 &0.390 &0.337 &0.382 &0.337 &0.263 &0.326 &0.344 \\
\hline
f&LLM-F &Starling &0.345 &0.355 &0.387 &0.520 &0.471 &0.396 &0.306 &0.457 &0.405 \\
g&&Llama &0.345 &0.227 &0.322 &0.426 &0.428 &0.316 &0.301 &0.438 &0.350 \\
h&&Gemma &0.329 &0.546 &0.456 &0.533 &0.525 &0.412 &0.429 &0.546 &0.472 \\
\hline\hline
i&\sys &Starling &0.518 &0.593 &\textbf{0.612}$^\bullet$ &\textbf{0.624}$^\bullet$ &\textbf{0.668}$^\bullet$ &\textbf{0.608}$^\bullet$ &$0.501^\bullet$ &\textbf{0.636}$^\bullet$ &\textbf{0.595}$^\bullet$ \\
j&&Llama &$0.522^\bullet$ &0.579 &$0.587^\bullet$ &0.572 &0.600 &0.541 &$0.497^\bullet$ &$0.631^\bullet$ &0.566 \\
k&&Gemma &\textbf{0.547}$^\bullet$ &\textbf{0.622}$^\bullet$ &\textbf{0.612}$^\bullet$ &0.599 &0.600 &0.521 &\textbf{0.556}$^\bullet$ &$0.632^\bullet$ & $0.586^\bullet$ \\
\hline
\end{tabular}
\caption{QWK Performance of \sys{} with the different LLMs compared to the baselines. 
\textbf{Bold} values indicate the best performance per trait, while values with $^\bullet$ outperform the SOTA.}\label{tab:main_results}
\end{table*}

We first examine the effectiveness of models trained only on the LLM-based features (LLM-F) without the other feature categories. The LLM-F models have the same architecture and training setup as \sys{}. Table \ref{tab:main_results} (rows f-h) presents their QWK performance. The results indicate that the extracted features demonstrate strong predictive capabilities across different LLMs. Notably, the LLM-F models clearly outperform the LLM direct scoring (LLM-D) (Table \ref{tab:main_results}, rows c-e) by an average of 9 points. This highlights that LLMs struggle with zero-shot essay scoring, which is consistent with previous findings~\cite{mansour-etal-2024-large}.

Among the three LLMs, Llama exhibits the lowest performance, followed by Starling, while Gemma
excels. 
Gemma generates the fewest questions, averaging 8.6 features per trait. This suggests that its generated features are concise and better aligned with the actual criteria in the rubrics compared to the other LLMs. In contrast, Llama generates an average of 20 features per trait; yet, its performance lags behind the others, indicating that some of the generated features introduce noise and do not align well with the scoring criteria.  

Remarkably, the ORG and CNV traits were the most difficult to score. In ASAP, these traits include prompts with different types and grade levels, making it more challenging to capture the characteristics of different prompts. Conversely, the CNT and PA traits achieved the highest QWK scores. Content-based traits are known to be difficult to predict, as they cannot be assessed heuristically, and the model must understand the prompt and the content of the essay in order to score it effectively \cite{li-ng-2024-automated}. Nevertheless, the features derived from LLMs alone were able to achieve a reasonable QWK, underscoring the predictive capabilities inherent in these features.

Although LLM-F models exhibit significantly lower performance than the baselines, the extracted features show good predictive capability for all traits (reflected in \emph{positive} QWK values). 
However, the results indicate that the trait-specific features are \emph{not sufficient} for effective essay trait scoring, suggesting the necessity of integrating other features for a comprehensive scoring framework.

\subsection{Effectiveness of \sys{} (RQ2)} 
\label{sec:RQ2}
Table \ref{tab:main_results} (rows i-k) shows the performance of \sys{} (combining all types of features) with the three LLMs. Notably, \sys{} outperforms the SOTA models on average over all traits using Starling and Gemma by 2 and 1 points, respectively, while it trails by 1 point with Llama. In terms of individual traits, Starling outperforms SOTA in 6 out of 8 traits (with on-par performance for the remaining two)
, Gemma in 5, and Llama in 4. More importantly, \sys{} establishes new SOTA performance for all traits on the ASAP dataset. 

Interestingly, while Gemma demonstrated the best performance with LLM-F, it did not achieve the same with \sys. This suggests its extracted features overlap with other features, leading to less improvement when combined. In contrast, Starling features perform better when integrated with other features, suggesting their extracted features are more unique and complementary. 

We note that ORG, CNV, and CNT traits showed improvements with the three LLMs, with an average of 1.1, 6.8, and 3.7 points, respectively.
Besides the positive impact of trait-specific features, we believe that the added prompt-specific features play a significant role in this improvement. These features are particularly applicable to those traits because the prompts within those traits exhibit the greatest diversity in terms of grade levels and essay types.

In contrast, PA, NAR, and LNG traits showed less improvements, with only Starling outperforming the SOTA. It is important to note that these traits already had good QWK scores with the feature-based baseline (Li \& Ng), indicating the effectiveness of the generic features. 
Although \sys{} did not outperform on those traits with Llama and Gemma, \sys{} with Starling showed improvements of 0.5, 3, and 1.20 points, respectively. This emphasizes the significant influence of the choice of LLM on the trait scoring performance.



\subsection{Feature Category Ablation Study (RQ3)} \label{sec:RQ3}



\begin{table*}[htp]\centering
\setlength{\tabcolsep}{5pt}
\normalsize
\begin{tabular}{ll|rrrrrrrr|r}
\hline
\textbf{Category} &\textbf{Size} &\textbf{ORG} &\textbf{WC} &\textbf{SF} &\textbf{PA} &\textbf{NAR} &\textbf{LNG} &\textbf{CNV} &\textbf{CNT} &\textbf{Avg}\\
\hline
Trait-specific &18.2 &2.23 &\textbf{3.73} &\textbf{2.00} &\textbf{10.22} &\textbf{12.27} &\textbf{13.28} &\textbf{8.73} &\textbf{8.35} &\textbf{7.60} \\
Prompt-specific &4 &\ul{4.57} &-3.35 &-1.69 &\ul{4.76} &\ul{8.94} &3.28 &3.34 &\ul{5.28} &\ul{3.14} \\
Length-based &16 &3.39 &\ul{1.53} &-0.33 &-0.70 &4.57 &2.94 &\ul{3.49} &3.42 &2.29 \\
Readability &12 &0.97 &-1.43 &-2.63 &2.26 &1.49 &\ul{7.39} &2.83 &2.58 &1.68 \\
Text complexity &5 &1.17 &-1.09 &-0.20 &2.19 &5.59 &5.82 &-1.62 &2.47 &1.79 \\
Text variations &43 &\textbf{7.27} &-0.70 &-0.95 &1.06 &3.59 &-0.44 &3.46 &0.10 &1.67 \\
Sentiment &5 &2.01 &-1.75 &-0.58 &0.64 &3.45 &4.71 &1.01 &0.23 &1.22 \\
\hline
\end{tabular}
\caption{\textbf{Drop} in QWK performance of \sys{} with Starling on ASAP when excluding one feature category. \textbf{Bold} and \ul{underlined} values indicate the most and second-most important categories for each trait, respectively.
}\label{tab:ablation_study}
\end{table*}

In this section, we address an important question regarding the significance of feature categories within our framework. 
This analysis helps in understanding whether each category is contributing positively to the prediction of trait scores. We conducted this experiment for \sys{} with Starling. Table~\ref{tab:ablation_study} presents the \emph{drop} in QWK when the regression model is trained with all feature categories but one. 

The trait-specific features stand out as the most significant category across all traits except organization. Notably, these are the only features that vary between traits, underscoring their critical role. This highlights the importance of incorporating \textit{trait-specific} features to effectively capture the unique requirements of each trait. It is also important to note that these are the only features that are generated automatically, whereas all other features require manual feature engineering.

The prompt-specific category comes next by positively contributing to 6 traits. For WC and SF, the negative impact can be attributed to the limited diversity of prompts within those traits. 
Within the generic features, the length-based category positively contributed to 6 traits; 
it contains a diverse range of features that pertain to various essay aspects, such as words and sentences, which may explain the negative values observed in some traits, as not all features in the category are equally relevant to every trait. 
Additionally, readability features had the greatest impact on LNG trait, which is expected as it emphasizes grammar and spelling accuracy.

The results indicate that all feature categories contribute positively on average. This also supports our proposed approach, where instead of relying exclusively on an LLM-based system, we leverage the classic feature categories. Moreover, the study clearly underscores the necessity for developing trait-specific AES systems, as the significance of features varies across different traits, demonstrating that not all features are relevant for every trait except for our trait-specific features.  


\subsection{Evaluation on ELLIPSE Dataset (RQ4)}
\label{sec:RQ4}

\begin{table*}[!htp]\centering
\begin{tabular}{lccccccc}
\hline
\textbf{Model} &\textbf{COH} &\textbf{SYN} &\textbf{VOC} &\textbf{GRM} &\textbf{CNV} &\textbf{PHR} &\textbf{Avg} \\
\hline
ProTACT$'$ &0.33 &0.35 &0.42 &0.29 &0.36 &0.36 &0.35 \\
\hline \hline
GP-F &0.45 &0.49 &0.48 &0.40 &0.50 &0.46 &0.46 \\
LLM-D &0.24 &0.29 &0.17 &0.28 &0.23 &0.20 &0.24 \\
LLM-F &0.35 &0.38 &0.38 &0.45 &0.44 &0.34 &0.39 \\
\hline \hline
\sys &\textbf{0.52} &\textbf{0.54} &\textbf{0.52} &\textbf{0.51} &\textbf{0.56} &\textbf{0.53} &\textbf{0.53} \\
\hline
\end{tabular}
\caption{QWK performance of \sys{} with Starling on ELLIPSE. \textbf{Bold}
values are the best per trait.
}
\label{tab:ellipse_results}
\end{table*}

Establishing a new SOTA performance on ASAP has pushed us to check whether \sys{} can exhibit similar performance on another dataset. We chose ELLIPSE because it contains several (44) prompts, making it well-suited for cross-prompt setup. Additionally, being a dataset for English learners allows us to test the applicability of \sys{} across different types of learners. It is worth mentioning that this is the \emph{first-ever} cross-prompt AES study on that dataset. 

To conduct this study, all the steps of \sys{} are repeated with Starling (the best-performing LLM on ASAP).
We adopted an 11-fold cross-validation approach, where each fold has 4 prompts, setting 1 unseen fold for testing.\footnote{\url{https://github.com/Sohaila-se/TRATES}}
The results are presented in Table \ref{tab:ellipse_results}.
We compare \sys{} with 4 baselines: ProTACT$'$ (a variant of ProTACT without the topic coherence (TC) features, as their extraction was not included in the provided implementation), LLM-D, LLM-F, and a feature-based model that is trained on the generic and prompt-specific features (GP-F). 

We first note that \sys{} clearly outperforms all the baselines in all traits, with a margin of at least 6.5 points on average. 
However, the performance of all models on ELLIPSE is lower than that on ASAP. This disparity can be attributed to several factors. First, the features utilized in ProTACT$'$ and GP-F were primarily developed on and tested for ASAP, raising concerns about their generalizability to other datasets, 
particularly for scoring essays written by English learners, whose writing styles and errors often differ from those of native speakers.
Nonetheless, GP-F outperforms LLM-F by 7 points,
likely due to the number of features; GP-F is trained on 89 features, whereas the average number of features for LLM-F from ELLIPSE is only 7. Additionally, the rubrics in ELLIPSE are more concise than those for ASAP, resulting in fewer generated features. The features generated from ASAP rubrics range from 5 to 13, while those from ELLIPSE rubrics range from 4 to 10. 


Finally, we recall that the results reflect the average score across 44 prompts (compared to 8 in ASAP), while QWK for individual prompts ranges from 0.25 to 0.87. It is also worth noting that assessing essays within this dataset is inherently difficult, as indicated by an inter-rater agreement between the human annotators, yielding a kappa value of less than 0.6 \cite{crossley2023english}, which is not far from the performance of \sys.

\subsection{Inference Cost} \label{sec:cost}

\begin{table*}[h]
    \centering
    \begin{tabular}{lcccccccc}
        \hline
         & \textbf{ORG} & \textbf{WC} & \textbf{SF} & \textbf{PA} & \textbf{NAR} & \textbf{LNG} & \textbf{CNV} & \textbf{CNT} \\ \hline
        Num. of Trait-specific Features & 25 & 9 & 13 & 18 & 20 & 17 & 15 & 31 \\
        Essay Average Length & 350 & 350 & 350 & 100 & 100 & 100 & 350 & 350 \\ \hline
        Trait-specific FE Time (msec)& 5,655 & 2,017 & 2,933 & 2,252 & 2,477 & 2,084 & 3,379 & 7,004 \\
        Generic + Prompt FE Time (msec) & 139 & 139 & 139 & 43 & 43 & 43 & 139 & 139 \\ 
        Regression Model Time (msec) & 0.12 & 0.12 & 0.12 & 0.12 & 0.12 & 0.12 & 0.12 & 0.12 \\ \hline
        Total Time (msec) & 5794 & 2156 & 3072 & 2295 & 2520 & 2127 & 3518 & 7143 \\ \hline
    \end{tabular}
    \caption{Average inference time of \sys{}, using the Starling LLM, per essay of prompt P1 for the ORG, WC, SF, CNV, and CNT traits, and P3 for the other traits. 'FE' denotes Feature Extraction stages. Time is in milliseconds.}
    \label{tab:time_comparison}
\end{table*}

In addition to \sys{} predictive performance, the practical deployment of \sys{} depends on its inference efficiency. We evaluate the inference cost of \sys{} in terms of processing time, focusing on the feature extraction stages and essay scoring. 

Table \ref{tab:time_comparison} presents the average inference time per essay for each trait with the Starling LLM. For the ORG, WC, SF, CNV, and CNT traits, we report the average inference time over the 1,783 essays of prompt P1. For PA, NAR, and LNG traits, the reported times are averaged over the 1,726 essays of prompt P3. All timings were obtained on an Azure VM equipped with an NVIDIA A10 GPU and an AMD EPYC 74F3 24-Core Processor. The model was loaded using the Hugging Face Transformers library with FP16 precision. 

As expected, the majority of the inference cost is attributed to LLM-based feature extraction, with times ranging from 2.02 to 7 seconds, depending on the number of features per trait. Traits from P3 show lower inference times compared to those from P1, primarily due to the shorter essay length, which reduces the LLM prompt size and speeds up processing. In contrast, the time for extracting the generic and prompt-specific features is 0.04 seconds for P3 and 0.14 seconds for P1, 
whereas the final scoring step using the neural network regression model is extremely fast and consistent across all traits, taking only 0.12 milliseconds. These results highlight that the LLM component highly dominates the overall inference cost.

These findings demonstrate the feasibility of deploying \sys{} in applications with moderate-latency requirements. Although latency remains a limitation and warrants further research, AES is not considered a real-time task and can tolerate the current processing delays.

\section{Conclusion and Future Work}
\label{sec:conclusion}

In this paper, we introduced \sys{}, a cross-prompt rubric-based framework utilizing LLMs, traditional features, and classical regression models for scoring essay traits. The framework is designed to reconceptualize the use of LLMs as feature generators and extractors. The results demonstrate the effectiveness of \sys{} yielding new SOTA performance on ASAP dataset and setting the first baseline for ELLIPSE dataset.
Our findings underscore the positive impact of the added LLM-based trait-specific features on enhancing the accuracy of trait scoring, emphasizing the necessity for novel models tailored to individual traits. 

In future work, we plan to extend the framework to holistic scoring and explore different prompting techniques for LLM-based trait-specific feature generation. Moreover, studying the usefulness of the generated sub-traits as feedback to the students remains an important direction for future work.

%
\section*{Acknowledgment}
This work was supported by NPRP grant\# NPRP14S-0402-210127 from the Qatar National Research Fund (a member of Qatar Foundation). The statements made herein are solely the responsibility of the authors.
\section{Limitations}

While \sys{} is advancing the use of LLMs for AES and extending their application beyond conversational tasks, some limitations are present in this study. Firstly, the study focused solely on relatively small LLMs, raising questions about whether larger ones would perform better in generating trait-specific questions and features. 

Furthermore, the performance of \sys{} remains unexplored in the context of holistic scoring, as holistic rubrics are often highly prompt-specific.

Moreover, our score-scaling approach was designed to address the discrepancies between the rubrics present in the ASAP dataset. The assigned scaling ranges were intuitively determined after a thorough examination of the various rubrics. It is debatable whether essays graded using different rubrics can be effectively combined to develop an AES system; the criteria for assessing essay quality may not be applicable across various rubrics, leading to inconsistencies in scores assigned to essays written for different prompts. Therefore, the research community should collaboratively establish a standardized mapping approach to align the rubrics of existing datasets within a unified framework. This would facilitate the creation of more realistic AES systems that can leverage diverse datasets for training and testing, 
addressing a wide range of writing proficiencies and real scenarios. 

Another limitation of this study is that the effectiveness of the generated questions as feedback for students was not assessed. However, the conceptual framework was discussed with experts in psychometrics and education fields, and the potential value of such feedback was acknowledged. We are dedicated to making these questions available to the community for future research. 
Moreover, the efficiency remains a limitation of our proposed framework and is an important area for future improvement.

Finally, the effectiveness of this work heavily relies on the quality of the grading rubric.


\bibliography{custom}

\appendix


\section{\sys{} LLM Prompts}

Our framework uses two LLM prompts: one for generating assessment questions based on the rubric (Figure \ref{fig:question_generation_prompt}) and another for extracting feature values by answering questions about the input essay (Figure \ref{fig:question_answering_prompt}).

\label{appendix:llms_prompts}
\begin{figure}[htp]
    \centering
    \begin{minipage}{\linewidth}
        \begin{tcolorbox}[colback=gray!5, colframe=black, boxrule=0.5pt, left=2pt, right=2pt, top=2pt, bottom=2pt]
{\fontsize{10}{10}\selectfont 
Your task is to formulate a set of assessment questions from the given rubric to be used to evaluate the <\textcolor{blue}{\textit{trait}}> of essays written by <\textcolor{blue}{\textit{grade-level range}}> grade students.

Here are some instructions to follow:\\
- Formulate the questions to rate the essay's aspects as High/Medium/Low \\
- The questions should start with "How would you rate ...".\\
- Keep the questions short and concise.\\
- Each question should address only one scoring criterion from the rubric.\\
- Structure your response in a numbered list from 1 to n, as follows: \\
1- <question 1?> \\
n- <question n?> 

---

Rubric: <\textcolor{blue}{\textit{rubric}}>

Questions: 

}
\end{tcolorbox}
        \captionof{figure}{LLM prompt for questions (i.e., features) generation.}
        \label{fig:question_generation_prompt}
    \end{minipage}
\end{figure}





\clearpage
\onecolumn
\LTcapwidth=\textwidth

\begin{longtable}
{p{0.17\textwidth}|p{0.23\textwidth}p{0.52\textwidth}}
\hline
\textbf{Category} & \textbf{Feature} & \textbf{Description} \\ \hline
\endfirsthead
\hline
\textbf{Category} & \textbf{Feature} & \textbf{Description} \\ \hline
\endhead
\endfoot

\endlastfoot

\hline
Length-based & mean\_word & Average word length. \\ 
& word\_var & Variance of word length. \\ 
& mean\_sent & Average sentence length. \\ 
& sent\_var & Variance of sentence length. \\ 
& ess\_char\_len & Total number of characters in the essay. \\ 
& word\_count & Total word count in the essay. \\ 
& prep\_comma & The number of prepositions and commas in the essay. \\ 
& characters\_per\_word & Average number of characters per word. \\ 
& syll\_per\_word & Average syllables per word. \\ 
& type\_token\_ratio & The ratio of unique words to the total number of words. \\ 
& syllables & Total syllable count. \\ 
& wordtypes & Total number of unique word types. \\ 
& sentences & Total number of sentences. \\ 
& long\_words & The total number of words with characters $\geq$ 7. \\ 
& complex\_words & The total number of complex words with syllable count $\geq$ 3. \\ 
& complex\_words\_dc & Complex words based on Dale-Chall readability formula. \\ 

\hline
Readability & spelling\_err & Count of spelling errors. \\ 
& automated\_readability & Calculated based on the average characters per word and average words per sentence. \\ 
& linsear\_write & Linsear Write readability score. \\ 
& Kincaid & Measures the grade level based on average words per sentence and average syllables per word. \\ 
& Coleman-Liau & Measures readability using average characters per 100 words and average sentences per 100 words. \\ 
& FleschReadingEase & Estimates reading ease based on average syllables per word and average words per sentence. \\ 
& GunninGP-FgIndex & Measures readability by analyzing average sentence length and the percentage of complex words. \\ 
& LIX & A readability measure based on sentence length and the number of long words. \\ 
& SMOGIndex & Index estimates the years of education needed to understand a text. \\ 
& RIX & Calculates text difficulty based on the number of long words and the number of sentences. \\ 
& DaleChallIndex & Uses a list of familiar words to calculate readability. \\ 

\hline
Text Complexity & clause\_per\_s & Average number of clauses per sentence. \\ 
& mean\_clause\_l & Average clause length. \\ 
& max\_clause\_in\_s & Maximum number of clauses in a sentence. \\ 
& sent\_ave\_depth & Average depth of sentence syntactic trees. \\ 
& ave\_leaf\_depth & Average depth of leaf nodes in syntactic trees. \\ 

\hline
Text Variations & unique\_word & Count of unique words. \\ 
& stop\_prop & Proportion of stop words. \\ 
& ",", "." &  Proportion of commas, and periods. \\ 
& VBG, VBZ, VBP, VB, VBD, VBN & Proportion of the different forms of verbs. \\ 
& NN, NNP, NNS & Proportion of the different forms of nouns. \\ 
& JJ, JJS, RBR, JJR & Proportion of the different forms of adjectives. \\ 
& RB, WRB & Proportion of the Adverbs and Wh-adverbs. \\ 
& PRP, WP, PRP\$ & Proportion of the different forms of pronouns. \\ 
& IN, MD, RP, CC, TO, WDT & Proportion of Other grammatical categories. \\ 
& DT, CD & Proportion of the numerals and determiners. \\ 
& POS & Proportion of genitive marker. \\ 
& tobeverb & Count of "to be" verbs. \\ 
& auxverb & Count of auxiliary verbs. \\ 
& conjunction & Count of conjunctions. \\ 
& pronoun & Count of pronouns. \\ 
& preposition & Count of prepositions. \\ 
& nominalization & Count of nominalized forms (e.g., "decision" from "decide"). \\ 
& begin\_w\_pronoun & The number of sentences that begin with a pronoun. \\ 
& begin\_w\_interrogative & The number of sentences that begin with an interrogative word. \\ 
& begin\_w\_article & The number of sentences that begin with an article. \\ 
& begin\_w\_subordination & The number of sentences that begin with a subordinating conjunction. \\ 
& begin\_w\_conjunction & The number of sentences that begin with a coordinating conjunction. \\ 
& begin\_w\_preposition & The number of sentences that begin with a preposition. \\ 

\hline
Sentiment & positive\_sentence\_prop & Proportion of positive sentences. \\ 
& negative\_sentence\_prop & Proportion of negative sentences. \\ 
& neutral\_sentence\_prop & Proportion of neutral sentences. \\ 
& overall\_positivity\_score & Overall positivity score based on sentiment analysis. \\ 
& overall\_negativity\_score & Overall negativity score based on sentiment analysis. \\ 
\hline
\caption{The list of the generic writing-quality features for the five categories.} \label{appendix:feature_list}
\end{longtable}

\begin{table*}[htp]
\centering

\begin{tblr}{
  row{even} = {c},
  rowsep=0pt,
  row{3} = {c},
  row{5} = {c},
  row{7} = {c},
  row{9} = {c},
  cell{1}{2} = {c},
  cell{1}{3} = {c},
  cell{1}{4} = {c},
  cell{1}{5} = {c},
  cell{1}{6} = {c},
  cell{1}{7} = {c},
  cell{1}{8} = {c},
  cell{1}{9} = {c},
  cell{1}{10} = {c},
  cell{1}{11} = {c},
  cell{1}{12} = {c},
  vline{2,5} = {-}{},
  hline{1-2,10} = {-}{},
}
Prompt & Scores & Ave Length & Essays & CNT                   & ORG              & WC              & SF          & CNV              & PA          & LNG                  & NAR                                 \\
P1    & 1 - 6         & 350        & 1783          &  $\checkmark$  &  $\checkmark$  &  $\checkmark$  &  $\checkmark$  &  $\checkmark$  &                           &                           &                                             \\
P2    & 1 - 6         & 350        & 1800          &  $\checkmark$  &  $\checkmark$  &  $\checkmark$  &  $\checkmark$  &  $\checkmark$  &                           &                           &                                                     \\
P3    & 0 - 3         & 100        & 1726          & $\checkmark$   &                           &                           &                           &                           &  $\checkmark$  &  $\checkmark$  &  $\checkmark$                             \\
P4    & 0 - 3         & 100        & 1772          &  $\checkmark$  &                           &                           &                           &                           & $\checkmark$ & $\checkmark$ & $\checkmark$                            \\
P5    & 0 - 4         & 125        & 1805          & $\checkmark$ &                           &                           &                           &                           & $\checkmark$ & $\checkmark$ & $\checkmark$                            \\
P6    & 0 - 4         & 150        & 1800          & $\checkmark$ &                           &                           &                           &                           & $\checkmark$ & $\checkmark$ & $\checkmark$                            \\
P7    & 0 - 3         & 300        & 1569          & $\checkmark$ & $\checkmark$ &                           &                           & $\checkmark$ &                           &                           &                                                      \\
P8    & 1 - 6        & 600        & 723           & $\checkmark$ & $\checkmark$ & $\checkmark$ & $\checkmark$ & $\checkmark$ &                           &                           &                         
\end{tblr}
\caption{A description of the ASAP and ASAP++ Datasets: Scores, Average essay length in terms of words, and Traits. 
}
\label{tab:dataset}
\end{table*}

\twocolumn

\begin{figure}[htp]
    \centering
    \begin{minipage}{\linewidth}
        \begin{tcolorbox}[colback=gray!5, colframe=black, boxrule=0.5pt, left=2pt, right=2pt, top=2pt, bottom=2pt]
{\fontsize{10}{10}\selectfont
You will be given a <\textcolor{blue}{\textit{essay\_type}}> essay written in response to the given prompt by a student in <\textcolor{blue}{\textit{grade\_level}}>th grade. Your task is to answer an assessment question with high/medium/low to evaluate the <\textcolor{blue}{\textit{trait}}> of the essay.\\
---

Follow the following format.\\
Prompt: the topic to which the essay responds.\\
Essay: the essay you need to evaluate.\\
Assessment Question: the question you need to answer about the essay.\\
Answer (High, Medium, or Low): your answer to the question.\\
---

Prompt: <\textcolor{blue}{\textit{task\_prompt}}>

Essay: <\textcolor{blue}{\textit{essay\_text}}>

Evaluation Question: <\textcolor{blue}{\textit{question}}>

Answer (High, Medium, or Low):

}
\end{tcolorbox}
        \captionof{figure}{LLM prompt for feature extraction.}
\label{fig:question_answering_prompt}
    \end{minipage}
\end{figure}

\begin{table}[h]
    \centering
    \begin{tabular}{l|c}
        \hline
        \textbf{Parameter} & \textbf{Value} \\ 
        \hline
        \#Essays & 6,482 \\ 
        Grade levels & 8–12 \\ 
        Number of prompts & 44 \\ 
        Average prompt size & 147 essays \\ 
        Average essay length & 427 words \\ 
        Score range & 1–5 (0.5 increments) \\ 
        \hline
    \end{tabular}
    \caption{ELLIPSE dataset statistics}
    \label{tab:dataset_ellipse}
\end{table}

\begin{table*}\centering
\begin{tabular}{llll}
\hline
\textbf{\#} &\textbf{Hyperparameter} &\textbf{Default value} &\textbf{Possible values} \\
\hline
1 &Loss function &MSE &MSE, Weighted-MSE \\
2 &Learning rate &0.001 &0.01,0.001,0.0001 \\
3 &Hidden layers &1 &1, 2, 3 \\
3 &Neurons per layer &32 &16, 32, \\
4 &Activation &ReLU &ReLU, SELU, LeakyReLU, Tanh, ELU \\
5 &L2 regularization &0 &0, 1e-6, 1e-5, 1e-4, 1e-3, 1e-2, 0.1 \\
6 &Dropout &0 &0.0, 0.1, 0.2, 0.3, 0.4, 0.5 \\
\hline
\end{tabular}
\caption{The hyperparameters search space. Parameters are listed in the order of tuning, along with their default values. The number of hidden layers and their sizes are tuned together.}
\label{tab:hyperparameters}
\end{table*}

\begin{table*}[htp]\centering
\begin{tabular}{lccccccccc}
\hline
&\textbf{ORG} &\textbf{WC} &\textbf{SF} &\textbf{PA} &\textbf{NAR} &\textbf{LNG} &\textbf{CNV} &\textbf{CNT} &\textbf{Avg} \\
\hline
Starling &25 &9 &13 &18 &20 &17 &15 &31 &18.5\\
Llama &30 &20 &10 &16 &16 &18 &20 &32 &20.3\\
Gemma &14 &8 &5 &6 &6 &6 &11 &13 &8.63\\
\hline
\end{tabular}
\caption{The number of LLM-generated questions from all the unique rubrics of each trait for the ASAP dataset.}
\label{tab:stat_question}
\end{table*}

\begin{table*}
\centering
\begin{tblr}{
  rowsep=0pt,
}
\hline
\textbf{Trait} &\textbf{Prompts} &\textbf{Example} \\
\hline
ORG &P1, P2 &Does the essay have a clear introduction, body, and conclusion? \\
&P7 &How well does the essay establish a clear thesis statement or central idea? \\
&P8 &{How would you rate the essay's transitions among sentences, paragraphs, \\ and ideas in terms of their smoothness and effectiveness?} \\
\hline
WC &P1, P2, P8 &{How would you rate the attempts at colorful language in the essay? \\ Are there any instances where the language seems overdone or forced?} \\
\hline
SF &P1, P2, P8 &{How well does the essay demonstrate variety and control in its sentence \\ structure, length, and beginnings?} \\
\hline
PA &P3, P4 &How would you rate the essay's adherence to the prompt's topic? \\
&P5, P6 &How would you rate the clarity of the essay's main points and arguments? \\
\hline
NAR &P3, P4 &How would you rate the essay's overall interest and engagement for the reader? \\
&P5, P6 &How would you rate the essay's overall interest and engagement? \\
\hline
LNG &P3, P4 &How would you rate the essay's vocabulary range and usage? \\
&P5, P6 &How would you rate the essay's grammar and spelling? \\
\hline
CNV &P1, P2, P8 &{How would you rate the essay's capitalization? Are there any significant errors \\ or inconsistencies?} \\
&P7 &How would you rate the essay's spelling accuracy, considering the grade level? \\
\hline
CNT &P1, P2, P8 &How well do the main ideas stand out in the essay? \\
&P3, P4 &How well does the essay maintain focus on the topic and avoid digressing? \\
&P5, P6 &How would you rate the essay's language and style? \\
&P7 &How would you rate the essay's language and tone? \\   
\hline
\end{tblr}

\caption{Examples of the assessment questions generated by Starling LLM for each unique rubric in the ASAP dataset. Each row represents a rubric in the dataset with the prompts associated with this rubric.}
\label{tab:question}
\end{table*}

\begin{table*}\centering
\begin{tblr}{
  rowsep=0pt,
  colspec = {l c l},
}
\hline
\textbf{Trait} &\textbf{\#Question}s &\textbf{Example} \\
\hline
Cohesion &5 &How would you rate the essay's overall cohesion and organization? \\
Syntax &4 &How would you rate the essay's use of punctuation and capitalization? \\
Vocabulary &6 &How would you rate the essay's control of word choice and word forms? \\
Grammar &10 &How would you rate the essay's use of punctuation? \\
Conventions &10 &How would you rate the essay's spelling accuracy? \\
Phraseology &5 &{How would you rate the essay's avoidance of noticeable repetitions \\ and misuses of phrases?} \\
\hline
\end{tblr}
\caption{Examples of the assessment questions generated by Starling LLM for each trait in the ELLIPSE dataset.}
\label{tab:question_ELLIPSE}
\end{table*}

\section{Generic Writing-Quality Features} \label{appendix:wq_features}

We used five categories of generic writing-quality features to cover different quality dimensions of the essay when scoring the different traits. The features categories are:
\begin{enumerate}
    \item \emph{Length-based} features, such as the number of words and sentences in the essay; these features are considered the most intuitive indicators for writing quality, which have been used extensively over the years \cite{mathias2018asap++,chen-li-2023-pmaes},
    \item \emph{Readability} features, which measure how difficult the essay is to read. This category includes readability scores, such as the automated readability index and the Flesch–Kincaid test \cite{ke2019automated}, 
    \item \emph{Text variations} features covering the usage of part-of-speech tags (POS) and punctuation.
    \item \emph{Text complexity} features, which evaluate the structural complexity of essays by analyzing the number of clauses per sentence and the sentence depths.
    \item \emph{Sentiment} features assessing the tone of the essays, capturing the proportion of positive, negative, and neutral sentences.
\end{enumerate}
Table \ref{appendix:feature_list} presents all the considered features in the five categories with their description. 

\section{ASAP Dataset Statistics}
The ASAP++ comprises 12,978 essays written in English in response to 8 prompts. Each of these prompts has a different number of responses and a different score range. There are a total of 10 different traits; each is covered in a subset of the prompts, except the content trait which is covered in all prompts. It's important to note that the traits of voice and style are only addressed in one prompt, making them unsuitable for cross-prompt experiments. Table \ref{tab:dataset} presents the statistics and the details of the dataset. 

\section{ELLIPSE Dataset Statistics} \label{ellipse_dataset_desc}
The ELLIPSE dataset comprises 6,482 essays written in response to 44 different prompts. These essays are written by English Language Learners across grade levels 8 through 12. All essays are evaluated using a standardized rubric that assesses 6 key traits: cohesion, syntax, vocabulary, phraseology, grammar, and conventions, in addition to a holistic score. Table \ref{tab:dataset_ellipse} presents the statistics and details of this dataset.

\section{LLMs Selection} \label{appendix:llms_desc}
We selected three LLMs for our experiments:
\begin{enumerate}
    \item  \textbf{
Starling}-LM-7B-beta~\cite{starling2023}: a fine-tuned model based on Mistral-7B and Openchat-3.5 via Reinforcement Learning from AI Feedback. It uses the Starling-RM-34B reward model with PPO for policy optimization. It leverages the Nectar ranking dataset and an improved training pipeline.\footnote{\url{https://huggingface.co/Nexusflow/Starling-LM-7B-beta}}
\item \textbf{Llama}-3.1-8B-Instruct~\cite{touvron2023llama}: a fine-tuned model based on Llama-3 via supervised fine-tuning and Reinforcement Learning with Human Feedback. It is optimized for multilingual dialogue tasks and outperforms many open-source and closed models on industry benchmarks.\footnote{\url{https://huggingface.co/meta-llama/Llama-3.1-8B-Instruct}}
\item \textbf{Gemma}-2-9b-it-SimPO~\cite{meng2024simpo} a fine-tuned model based on gemma-2-9b-it via Simple Preference Optimization (SimPO). SimPO is an offline method for enhancing LLM training with preference optimization datasets.\footnote{\url{https://huggingface.co/princeton-nlp/gemma-2-9b-it-SimPO}}
\end{enumerate}

\section{Zero-shot LLM Scoring} \label{appendix:zs_llm_scoring}

\begin{figure}[htp]
    \centering
    \begin{minipage}{\linewidth}
        \begin{tcolorbox}[colback=gray!5, colframe=black, boxrule=0.5pt, left=2pt, right=2pt, top=2pt, bottom=2pt]
{\fontsize{10}{10}\selectfont

You will be given a <\textcolor{blue}{\textit{essay\_type}}> essay written in response to the given prompt. Your task is to score the <\textcolor{blue}{\textit{trait}}> of the essay as per the given rubric.\\
---

Follow the following format.\\
Prompt: the topic to which the essay responds.\\
Rubric: the grading rubric to score the essay.\\
Essay: the essay you need to evaluate.\\
Score: the score of the essay as per the given rubric (only one number).\\
---

Prompt: <\textcolor{blue}{\textit{task\_prompt}}>

Rubric: <\textcolor{blue}{\textit{trait\_rubric}}>

Essay: <\textcolor{blue}{\textit{essay\_text}}>

Score:

}
\end{tcolorbox}
        \captionof{figure}{LLM prompt for Zero-shot LLM direct scoring (LLM-D).}
\label{fig:zs_llm_scoring}
    \end{minipage}
\end{figure}

The primary purpose of including this baseline is to demonstrate the performance of LLMs in AES, highlighting that direct LLMs utilization lags significantly behind other established methods in the field. The results of this baseline underscore the necessity for alternative ways to integrate LLMs in AES, rather than relying on them solely for scoring purposes. Figure \ref{fig:zs_llm_scoring} illustrates the LLM prompt used for zero-shot essay scoring.

\section{Hyperparameters}

The hyperparameters are tuned using sequential tuning with QWK as the scoring function. For each fold, the best hyperparameters on the validation set are used to train the regression model and evaluate its performance on the test set. The hyperparameters search space is presented in Table \ref{tab:hyperparameters}.

\section{Rubric-based Assessment Questions}
\label{sec:questions_appendix}

Table \ref{tab:stat_question} shows the total number of generated questions by each LLM from the rubrics of each trait. 
Additionally, Table \ref{tab:question} presents examples of the assessment questions generated by the Starling LLM for each rubric. These rubrics are derived from the ASAP and ASAP++ datasets \cite{mathias2018asap++}, which were utilized in our evaluation. 
The data indicates that Llama tends to generate the most numerous questions, followed by Starling and then Gemma.

For ELLIPSE dataset, one rubric for each trait is used to score all the essays in the dataset. Table \ref{tab:question_ELLIPSE} illustrates the number of generated questions from each rubric, and an example of the generated questions for each trait.


\section{Score-scaling} \label{appendix:score-scaling}


The rationale for scaling essay scores based on grade level stems from the fact that writing expectations are generally lower in the earlier grades compared to higher ones. To develop an AES system capable of accommodating all educational levels, we propose scaling the scores for each grade level within a defined range during the training process. This range is then converted back to the original scoring range for evaluation. We assigned the following score ranges to the different grade levels in the ASAP dataset: [0,4] for grade 7, [0,5] for grade 8, and [0,6] for grade 10,  
showing that the maximum score is decreased by one score level (1 point) for each lower grade relative to the maximum grade level within the dataset. Note that this is applicable only when the scoring rubrics for the different prompts in the dataset are different. Hence, we applied score-scaling only to the ASAP dataset.




\end{document}